# Language-Specific Sentiment Polarity Biases in Encoder and Large Language Model Classification of Product Reviews

**Advita Rajiv [1], Kavitha Kothur [2] and Gautham Reddy [3]**
[1] John P. Stevens High School, Edison, NJ 08820
[2] Ford Foundation, New York, NY 10017
[3] NC State University, Raleigh, NC 27606

**OVERVIEW:** This study investigates sentiment polarity biases, specifically, differences in how accurately AI models classify positive versus negative reviews across languages and model architectures. Large language models show a negative bias in French and are more accurate on negative reviews while encoder models exhibit positive bias in Japanese missing negative reviews that use indirect criticism. These language-specific polarity biases have implications in both social and business domains deploying multilingual sentiment analysis systems.

**SUMMARY**

Sentiment analysis systems are widely used to evaluate opinions across languages, but their reliability depends on their consistency regardless of sentiment type. We hypothesized that sentiment classification accuracy would differ between positive and negative reviews, and that these polarity biases would vary by language and model architecture. Using the Amazon Multilingual Reviews dataset (N = 500 per language), we tested this hypothesis by classifying French and Japanese product reviews with mDeBERTa-v3 (an encoder model) and three large language models (Claude Opus 4.6, GPT-5.4, Gemini 3.1 Pro). We found that all three LLMs showed negative bias in French, achieving 99.2% accuracy on negative reviews but a lower 91.6 to 94.4% on positive reviews ($p < 0.01$, Cramér's $V = 0.13–0.17$). Conversely, mDeBERTa showed positive bias in Japanese (92.4% positive vs. 84.4% negative, $p = 0.008$, $V = 0.12$) but no bias in French. These results support our hypothesis that polarity biases are both architecture-specific and language-specific.

## INTRODUCTION

In a globalized world, Natural language processing (NLP) systems play a key role in analyzing text across multiple languages. The Transformer architecture revolutionized NLP with its self-attention mechanism, enabling the model to learn contextual relationships between words regardless of their distance in text. Its ability to perform parallel processing reduced time to train and opened up the possibility of ingesting large amounts of information. (1). Two Transformer architectures overperformed previous state-of-the-art versions. Encoder-only models such as BERT (Bidirectional Encoder Representations from Transformers) and DeBERTa use the encoder to understand the context of words by looking at both the left and right sides of a token simultaneously. GPT (Generative Pre-trained Transformer) uses only the decoder and generates text one word at a time based on preceding context (2).

Sentiment analysis extracts opinions and emotions in written text such as product reviews, comments and blogs. As opinions are key influencers of human behavior, sentiment analysis is a widely applied NLP task in areas such as market research, social media monitoring and customer feedback analysis across business and social domains (3). Cross-lingual sentiment analysis extends this task across multiple languages whereby training in one language enables the model to perform the task in another language. However, research has shown that models do not perform equally well across all languages (4). This has an impact on the reliability of multilingual systems in global applications.

Previous studies have compared model architectures on overall accuracy across languages (4, 5). However, less attention has been paid to whether models show biases toward particular sentiment classes, specifically if they classify positive reviews differently from negative reviews. There has been recent research that has highlighted polarity biases in LLMs where GPT-3.5 tended to overclassify negative reviews as positive, while GPT-4 demonstrated the opposite (6). However, it is not clear if these biases vary by language or factor in cultural communication styles. Such biases could introduce errors in real-world applications where if a model is more accurate on negative reviews than positive ones, analyses and decisions based on that model would under count positive sentiment.

We studied sentiment polarity biases across two typologically distinct languages, French and Japanese. We selected these languages because they represent different positions on the cultural communication spectrum. Foundational work on high-context versus low-context cultures has established that Japanese communication leans on implicit, indirect messaging where meaning is conveyed through rhythm and context rather than explicit words (7). Subsequent research has shown that this style extends from face-to-face interactions to written contexts including online reviews, where Japanese writers tend to use more indirect criticism strategies (8). Although French is classified as moderately high-context within Europe, French communicators tend to express negative feedback more directly and with less mitigation than many high-context cultures (9). Research on sentiment analysis of French movie reviews has noted that rhetorical features including irony and mixed sentiment challenge automated classification (10). These differences in sentiment expression conventions among French and Japanese could create varying challenges for different model architectures, making these languages well suited for testing whether polarity biases vary by language.

We hypothesized that sentiment classification accuracy would differ between positive and negative reviews, with these polarity biases varying by model architecture and language. We studied two specific predictions. One, that encoder-based BERT and LLM architectures would show distinct polarity bias patterns (H1). Two, that these biases would manifest differently in French vs. Japanese reflecting the differences in sentiment expression (H2).

## RESULTS

### *Overall Accuracy*

We began by assessing each model's overall classification performance with a measure of accuracy levels. All four models achieved high overall accuracy in both languages. mDeBERTa achieved 92.2% accuracy on French and 88.4% on Japanese. The three LLMs performed comparably. Claude performed at 96.4% accuracy on French and 96.0% on Japanese, GPT-5.4 displayed an accuracy of 95.4% on French and 96.0% on Japanese, while Gemini had an accuracy of 96.8% on French and 96.2% on Japanese. Cross-lingual performance gaps were small in all models, ranging from 0.4 to 3.8 percentage points.

### *Sentiment Polarity Bias in French*

We compared each model's performance on positive versus negative French reviews using chi-square tests of independence. We observed polarity biases that differed by model architecture which was consistent with our hypothesis H1 (Table 1, Figure 1). The three LLMs classified negative reviews more accurately than positive reviews exhibiting statistically significant negative bias in French (all $p < 0.05$, $V > 0.10$). Claude classified 248 of 250 (99.2%) negative reviews accurately versus 234 of 250 (93.6%) positive reviews. GPT-5.4 showed a wider gap with 248 of 250 negative reviews (99.2%) vs. 229 of 250 positive reviews (91.6%) and Gemini followed a similar pattern with 248 of 250 negative (99.2%) vs. 236 of 250 positive reviews (94.4%). Effect sizes were small ($V = 0.13$ to $0.17$) but consistent across all three LLMs.

mDeBERTa, however, showed no significant difference in its accuracy predicting the sentiment of 231 of 250 negative reviews correctly (92.4%) vs. 230 of 250 positive reviews (92.0%).

| Model | Positive Accuracy | Negative Accuracy | Bias | Chi-Square $\chi^2$ | p-value | Cramér's *V* |
|---|---|---|---|---|---|---|
| mDeBERT a-v3 | 92.0% | 92.4% | -0.4pp | 0.00 | 1.00 | <0.01 |
| Claude Opus 4.6 | 93.6% | 99.2% | -5.6pp | 9.74 | 0.002** | 0.14 |
| GPT-5.4 | 91.6% | 99.2% | -7.6pp | 14.77 | <0.001*** | 0.17 |
| Gemini 3.1 Pro | 94.4% | 99.2% | -4.8pp | 7.81 | 0.005** | 0.13 |

**Table 1: Sentiment classification accuracy by polarity for French reviews.** Accuracy of positive and negative reviews (N = 250 per class) for the four models. Bias is calculated as positive accuracy minus negative accuracy. Negative bias indicates better performance on
negative reviews. Chi-square tests of independence with Yates' continuity correction were used to test for the model's performance differences between positive and negative reviews. All three LLMs showed statistically significant negative polarity bias (**p < 0.01, ***p<0.001)) with small magnitude of association (V 0.13 to 0.17). pp = percentage points.

*Sentiment Polarity Bias in Japanese*

We did a similar comparison for Japanese reviews to check if the patterns would hold. Instead, we noticed different results with mDeBERTa showing positive polarity bias. This was in line with our hypothesis H2 about the bias patterns being language-specific (Table 2, Figure 1).
mDeBERTa analyzed 231 of the 250 positive reviews correctly at a 92.4% accuracy compared to 211 of 250 negative ones at 84.4% accuracy. It showed statistically significant positive bias ($p = 0.008$, $V = 0.12$) and struggled with negative sentiment in Japanese reviews. The LLMs, however, showed comparable accuracy on Japanese positive and negative reviews (all $p > 0.05$, $V < 0.10$). Claude accurately predicted 238 of 250 positive reviews and 242 of 250 negative reviews (95.2% vs. 96.8%), GPT-5.4 did 236 positive and 244 negative (94.4% vs. 97.6%) and Gemini 239 positive and 242 negative (95.6% vs. 96.8%).

| Model | Positive Accuracy | Negative Accuracy | Bias | Chi-Square $\chi^2$ | p-value | Cramér's *V* |
|---|---|---|---|---|---|---|
| mDeBERT a-v3 | 92.4% | 84.4% | +8.0pp | 7.04 | 0.008** | 0.12 |
| Claude Opus 4.6 | 95.2% | 96.8% | -1.6pp | 0.47 | 0.49 | 0.03 |
| GPT-5.4 | 94.4% | 97.6% | -3.2pp | 2.55 | 0.11 | 0.07 |
| Gemini 3.1 Pro | 95.6% | 96.8% | -1.2pp | 0.22 | 0.64 | 0.02 |

**Table 2: Sentiment classification accuracy by polarity for Japanese reviews.** Accuracy on positive and negative Japanese reviews (N = 250 per class) for the four models. Bias is calculated as positive accuracy minus negative accuracy. Positive bias indicates better performance on positive reviews. Chi-square tests of independence with Yates' continuity correction were used to test for the model's performance differences between positive and negative reviews. mDeBERTa-v3 showed statistically significant positive polarity bias (** $p<0.01$) with small magnitude of association ($V=0.12$). pp = percentage points.

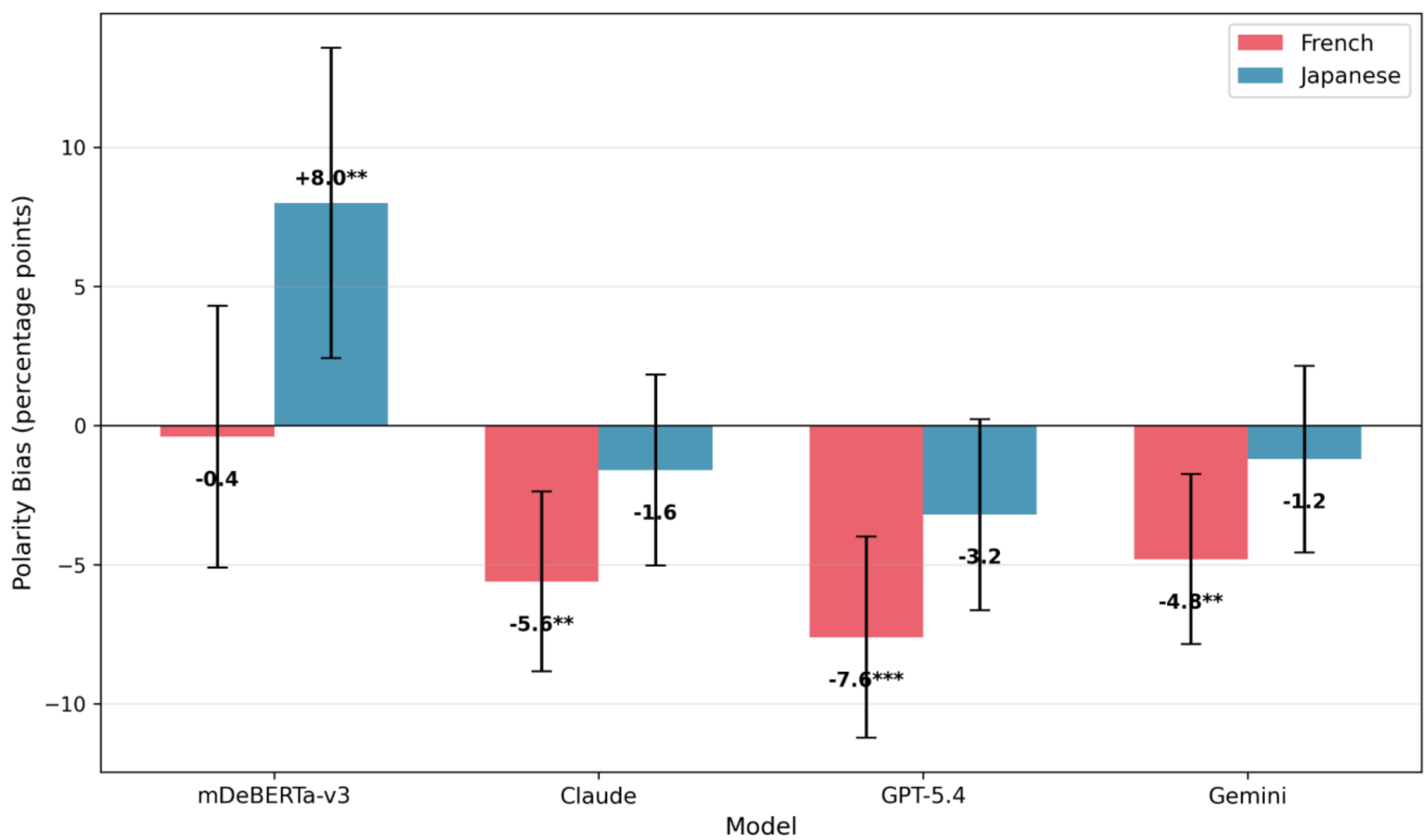

**Figure 1. Sentiment polarity bias by model and language.** Bars show the difference between positive and negative review accuracy for each model in French and Japanese. Negative values indicate the model performed better on negative reviews. Positive values indicate better performance on positive reviews. Error bars show 95% confidence intervals calculated using standard error for the difference between positive and negative review accuracy. mDeBERTa-v3 showed significant positive bias in Japanese, while LLMs showed significant negative bias in French. **p<0.01, ***p<0.001

*Error Type Analysis*

We studied the confusion matrices for each model-language combination in order to further understand the polarity biases. This distinguished between false negatives (positive reviews misclassified as negative) and false positives (negative reviews misclassified as positive), revealing the direction of each model's errors (**Table 3**).

In French, GPT-5.4 produced more false negatives than false positives (21 vs. 2) and with a 10:1 ratio it was 10 times more likely to misclassify a positive review as negative than vice versa. Claude and Gemini (16 vs. 2, 14 vs. 2) were 7 times more likely to misclassify a positive review as negative than vice versa. This explained the significant negative polarity bias observed earlier (**Table 1, Figure 1**). In contrast, mDeBERTa's error distribution had a similar spread across false negatives and false positives (20 vs. 19).

In Japanese, false positives exceeded false negatives in the case of mDeBERTa (39 vs. 19) with a 2:1 ratio confirming the positive polarity bias identified earlier (**Table 2, Figure 1**). The three LLMs maintained a balanced error distribution in Japanese with false positive and false negative counts remaining similar.

| Language | Model | False Negatives | False Positives | Error Direction |
|---|---|---|---|---|
| French | mDeBERTa-v3 | 20 (8.0%) | 19 (7.6%) | Balanced |
| | Claude Opus 4.6 | 16 (6.4%) | 2 (0.8%) | False Negative |
| | GPT-5.4 | 21 (8.4%) | 2 (0.8%) | False Negative |
| | Gemini 3.1 Pro | 14 (5.6%) | 2 (0.8%) | False Negative |
| Japanese | mDeBERTa-v3 | 19 (7.6%) | 39 (15.6%) | False Positive |
| | Claude Opus 4.6 | 12 (4.8%) | 8 (3.2%) | Balanced |
| | GPT-5.4 | 14 (5.6%) | 6 (2.4%) | Balanced |
| | Gemini 3.1 Pro | 11 (4.4%) | 8 (3.2%) | Balanced |

**Table 3. Distribution of error types by model and language.** Classification errors broken down into False Negatives (positive reviews misclassified as negative) and False Positives (negative reviews misclassified as positive). Percentages indicate error rate relative to true class size (N=250 reviews per sentiment class per language). In French, the 3 LLMs showed a directional bias towards False Negatives. In Japanese, mDeBERTa-v3 showed a directional bias towards False Positives.

*Inter-Model Agreement and Error Overlap*

We compared two models at a time and noted down their agreement rates and error overlap. This assessment tested if the two architectures processed sentiment classification differently (H1).

The three LLMs showed high levels of agreement on correct results ranging from 98.2% to 98.5%. This suggested they processed sentiment similarly most likely due to similar training data, architectures, and optimization techniques. In contrast, mDeBERTa agreed with each LLM on correct results in 90.6 to 91.4% of cases.

We analyzed the classification error set to evaluate whether they failed on the same or different sets of reviews. The error overlap between mDeBERTa and LLMs was low at 23.7 to 24.7%. This supported H1 with an indication that the encoder-based architecture failed on different types of reviews than the LLM architecture. The low agreement on errors shows that the architectures are complementary and can be combined into an Oracle ensemble (correct if any model is correct), using methods such as stacking or routing via logistic regression. The ensemble would have achieved 98.0% accuracy on both languages compared to 96.2 to 96.8% for the best single model.

*Patterns in Misclassified Reviews*

To further analyze the polarity biases, we performed a qualitative analysis of all misclassified reviews. We manually reviewed each error and noted down language features that might explain the classification errors.

For French, we analyzed the 14 to 21 positive reviews that each LLM incorrectly classified as negative. We found that 67 to 79% of these errors contained negative markers such as "mais" (but), "problème" (problem), or "malheureusement" (unfortunately). These were mixed-sentiment reviews where the reviewer expressed overall satisfaction while mentioning specific issues confusing the LLMs.

For Japanese, we analyzed the 39 negative reviews that mDeBERTa incorrectly classified as positive. We found that 97% contained indirect expression or positive framing while expressing dissatisfaction. However, LLMs correctly classified 82 to 85% of these same reviews that mDeBERTa missed, suggesting that LLMs were better able to parse negative sentiment embedded within polite or softened language.

## DISCUSSION

The two hypotheses we stated were supported by the results and the analysis we performed. Sentiment polarity biases differed between encoder and LLM architectures (H1) and manifested differently across French and Japanese (H2). Effect sizes were small (V = 0.12 to 0.17) but statistically significant.

In French, LLMs showed strong negative bias (better on negative reviews) while the encoder showed no bias. In Japanese, this pattern reversed with the encoder showing strong positive bias (worse on negative reviews) while LLMs showed no bias. This suggested that the polarity biases were not inherent to the architecture alone but to a combination of architecture and language.

For LLMs, the negative bias in French is a reflection of the challenges inherent to the style of critique in product reviews. As shown in our error analysis, LLMs appeared to over-weight explicit negative words despite the overall positive sentiment overriding broader contextual clues. Researchers have documented that textual data characteristics and linguistic features influence LLMs (6) and our findings aligned with it.

For mDeBERTa, the positive bias in Japanese emphasized the challenges of indirect sentiment expression. For example, reviewers began with phrases like "気に入っていたのに" (I liked it, but...) or "デザインはいいけど" (The design is good, but...) before describing product failures. This indirect style is characteristic of Japanese high-context communication, where negative feedback is typically softened to maintain group harmony (7). The NLI-based zero-shot classification approach used by mDeBERTa determines labels by measuring textual entailment between the review and hypothesis templates such as 'The sentiment is negative' (11). When negative sentiment is expressed indirectly, through softening language or positive framing, the review provides weaker lexical support for the negative hypothesis, likely causing the model to default to a positive prediction. As noted in our error analysis, LLMs were able to comprehend implicit negative sentiment and therefore correctly classified a majority of the reviews missed by the encoder model. Prior research has shown that adding cultural context improves classification of Asian language expressions and our LLM results aligned with it (12).

The high similarity among LLMs reflects similar design choices including transformer architectures, training dataset (web text, books, code) and optimization techniques such as fine-tuning and reinforcement learning from human feedback (13). This similarity implies they potentially have similar weaknesses such as if one LLM exhibits a bias, others likely will too.

The low error overlap between mDeBERTa and LLMs confirms that these architectures failed on different types of reviews therefore lending support for an ensemble approach. Future research should explore and test such approaches.

For those implementing multilingual sentiment analysis systems in global applications, there are practical implications to consider. A system using only LLMs for French analysis would undercount positive sentiment by 5 to 8 percentage points, potentially introducing bias in business decisions. Similarly, a system using only mDeBERTa for Japanese would miss negative reviews therefore underestimating, for example, customer complaints. When selecting models for real-world deployment, they should be evaluated for polarity specific accuracy besides overall accuracy.

Our study had several limitations. First, we examined only two languages and analyzed the patterns. The patterns may differ for other language pairs. Based on the Japanese results, we would predict that mDeBERTa would also show positive bias in other Asian languages from collectivistic, high-context cultures such as Korean and Chinese (14). However, because the communication styles vary across Asian cultures, this prediction requires empirical verification.

Second, our study used product reviews. The patterns for other domains such as social media and news may be different. Further research in these domains will validate our findings. Third, our sample size (500 per language) was adequate for the current study but using larger samples in future research would improve the precision of results. Fourth, the qualitative analysis of error patterns was conducted by us and lacked a formal assessment. Future work should factor in deeper analysis.

In conclusion, our findings demonstrated that sentiment polarity biases are both architecture-specific and language-specific. To prevent significant biases from affecting real-world applications, a thorough evaluation of a multilingual sentiment system should include not only overall accuracy but also performance on each sentiment class separately.

## MATERIALS AND METHODS

### *Dataset*

We extracted French and Japanese anonymized product reviews rated 1-5 stars from the publicly available Multilingual Amazon Reviews Corpus (15). By sourcing both the French and Japanese data from the same collection, we avoided domain specific confounding variables such as when comparing French movie reviews and Japanese reviews of consumer goods. We labeled 1–2 star reviews as negative and 4-5 star reviews as positive excluding 3 star reviews. We accessed it via the Hugging Face Datasets library.

### *Balanced Sampling*

We balanced the sample with 250 positive and 250 negative reviews from each language's test set using a fixed random seed (seed=42) yielding N=500 per language. The resultant sample set had comparable document lengths across languages with a mean token length of 55.8 for French (median: 45, range: 9–489) and 57.3 for Japanese (median: 41, range: 13–527). The minimum French token count of 9 corresponded to very brief reviews such as "superbe / pour des photos magnifiques" (superb / for magnificent photos).

### *Models*

The encoder model used was mDeBERTa-v3-base (MoritzLaurer/mDeBERTa-v3-base-mnli-xnli), a multilingual architecture fine-tuned on the MultiNLI and XNLI natural language inference datasets (16, 17). This model was run on CPU using the Hugging Face Transformers library (version 5.0.0) with PyTorch 2.10.0. The LLMs were Claude Opus 4.6 from Anthropic, GPT-5.4 from OpenAI, and Gemini 3.1 Pro from Google, accessed via their Python SDKs (anthropic v0.84.0, openai v2.26.0, google-genai v1.66.0) in February 2026. We used the standard provider defaults for all models. This included a temperature and top-p of 1.0, standard reasoning modes, and default token limits.

*Classification Procedure*

For mDeBERTa, we used zero-shot classification via the Hugging Face pipeline based on natural language inference (NLI). Each review was evaluated against the hypothesis template "The sentiment is {label}" for both "positive" and "negative" and the label with higher score was selected as the prediction. For LLMs, each review was submitted individually with the prompt: "Classify the sentiment of the following review as either 'positive' or 'negative'. Respond with a single word only." Responses were then parsed to extract the predicted label.

*Statistical Analysis and Error Coding*

Chi-square tests of independence with Yates' continuity correction were used to test whether accuracy differed between positive and negative reviews within each model-language combination. Effect sizes were calculated using Cramér's *V*, with values interpreted as negligible (<0.10), small (0.10–0.20), medium (0.20–0.30), or large (>0.30) following standard conventions (18). Confidence intervals for accuracy differences were calculated at 95% using the standard error for the difference between two proportions. Statistical significance was assessed at $\alpha = 0.05$. All statistical analyses were conducted in Python 3.12.12 using Scipy.stats version 1.16.3 (19). We manually reviewed all misclassified false positive and false negative predictions for each model-language combination to spot patterns. French reviews were coded for the presence of explicit negative markers (e.g., "mais," "problème," "malheureusement," "dommage," "contre"). Japanese reviews were coded for indirect constructions and positive framing that embedded negative sentiment. This qualitative analysis was exploratory rather than providing definitive explanations.

Code to support this study is available at the following GitHub link: [https://github.com/Statofthe7/SentimentAnalysisPolarityBias](https://github.com/Statofthe7/SentimentAnalysisPolarityBias).